\documentclass[letterpaper]{article}
\usepackage[margin=1in]{geometry}
\usepackage{graphicx,amsmath,amsthm,amssymb}
\usepackage{natbib}
\usepackage{url,color,booktabs}
\usepackage{hyperref}

\def \ifempty#1{\def\temp{#1} \ifx\temp\empty }

\newcommand{\w}{\ensuremath{\mathbf{w}}}
\newcommand{\x}{\ensuremath{\mathbf{x}}}
\newcommand{\y}{\ensuremath{\mathbf{y}}}

\newcommand{\btheta}{\ensuremath{\boldsymbol{\theta}}}

\newcommand{\bTheta}{\ensuremath{\boldsymbol{\Theta}}}

\newcommand{\bbR}{\ensuremath{\mathbb{R}}}

\newcommand{\norm}[2][]{%
  \ifempty{#1} {\left\lVert#2\right\rVert} \else {#1\lVert#2#1\rVert} \fi}

{%
\begin{list}{#1}{
\vspace{-\topsep}
\vspace{-\partopsep}
\setlength{\itemindent}{0cm}
\setlength{\rightmargin}{0cm}
\setlength{\listparindent}{0cm}
\settowidth{\labelwidth}{#1}
\setlength{\leftmargin}{\labelwidth}
\addtolength{\leftmargin}{\labelsep}
\setlength{\itemsep}{0cm}
}%
}%
{%
\end{list}
\vspace{-\topsep}
\vspace{-\partopsep}
}

{\begin{enumerate}%
}%
{\end{enumerate}}

\bibpunct[, ]{(}{)}{;}{a}{,}{,} 
					
\title{An Experimental Comparison \\of Old and New Decision Tree Algorithms}
\author{
    Arman Zharmagambetov \thanks{equal contribution} $\hspace{2ex}$ Suryabhan Singh Hada  \footnotemark[1]\\
    Miguel {\'A}.\ Carreira-Perpi{\~n}{\'a}n $\hspace{4ex}$ Magzhan Gabidolla\\
   Dept. of Computer Science and Engineering, University of California, Merced \\
   {\url{http://eecs.ucmerced.edu}}
	}
\date{March 20, 2020}

\definecolor{DarkRed}{rgb}{0.545,0,0}
\hypersetup{
  breaklinks=true, colorlinks=true, linkcolor=black,
  citecolor=black, urlcolor=DarkRed 
}
					
\begin{document}
					
\maketitle
\begin{abstract}
This paper presents a detailed comparison of a recently proposed algorithm for optimizing decision trees, tree alternating optimization (TAO), with other popular, established algorithms. We compare their performance on a number of classification and regression datasets of various complexity, different size and dimensionality, across different performance factors: accuracy and tree size (in terms of the number of leaves or the depth of the tree). We find that TAO achieves higher accuracy in nearly all datasets, often by a large margin.

\end{abstract}

\section{Introduction}
\label{intro}

Decision trees are one of the widely used statistical models. Apart from being a good classifier, they have some very unique properties which separate them from other models. A path from root to any leaf can be described as a sequence of decisions: `$x_i < b$' (axis-aligned trees) or $\w^T\x > b$ (oblique tree). This not only makes decision trees very fast at inference but also makes them good interpretable models. This sequence of decisions can be used as IF-THEN rules to understand the prediction of the model for a given input. 

However, there is one major drawback with decision trees: learning a tree from
data is a very difficult optimization problem, involving a search over a complex, large set of tree structures, and over the parameters at each node. In fact, it is well-known that the optimal binary tree construction problem is NP-hard \citep{HyafilRivest76a}. Recently, \cite{CarreirTavall18a} proposed Tree Alternating Optimization (TAO) algorithm to improve this problem, where authors directly optimize the misclassification error, using alternating optimization over separable subsets of nodes. In this work, we compare TAO against some well-known decision tree algorithms (both axis-aligned and oblique) over a wide range of classification and regression datasets. 

We have structured this paper in the following way: In section~\ref{s:algos} we briefly describe all the algorithms that we use for the comparison. Next, in section~\ref{s:datasets} we describe all the data sets including the number of instances and dimensionality for each data set. In section~\ref{s:expset} and~\ref{s:result}, we describe our experimental setup and results of the comparison. 

\section{The algorithms}
\label{s:algos}

The literature of tree learning both for classification and regression is immense. In this paper, we restrict our comparison to the most popular and established methods for training axis-aligned and oblique decision trees with constant leaves. Below we provide a short description of the algorithms which were used in our comparison. More details can be found on the corresponding cited papers. 
\begin{itemize}
\item \textbf{CART:} CART \citep{Breiman_84a} is one of the most widely used algorithms for training axis-aligned decision trees. It learns the tree by greedy recursive partitioning, to optimize the impurity measure at each node. At each growing stage for a given node, it enumerates through all the attributes to find the best split that reduces the Gini-index for that node. It grows the tree up to the max depth and then starts pruning nodes one by one such that it does not increase the misclassification error by a certain threshold.

\item \textbf{C5.0:} \cite{Quinlan93a} is known as an established univariate decision tree learning software. Similarly to CART, it uses a greedy recursive partitioning of the tree nodes. At each recursive split, the algorithm enumerates over different feature-threshold combinations and picks the best one according to the information gain criterion. Pruning can be applied once the tree growing phase is finished. We use this version of the decision tree learning algorithm developed by Ross Quinlan since it is the latest and extended version of the earlier ID3 and C4.5 algorithms \citep{Quinlan93a}.

\item \textbf{OC1:} Oblique Classifier 1 (OC1) \citep{Murthy_93a} is also based on the idea of greedily growing a large tree to minimize some impurity measure, and pruning it using the same method as in CART. However, it considers only oblique hyperplanes in internal nodes, where all the features are used in the decision making. To find the weights of the those nodes, it starts with a random guess, and iteratively perturbs one weight at a time until impurity measure does not improve. Then, the weights are changed in some random direction with the aim of moving out from a local minimum. The random movements that help to improve impurity are  further optimized using the iterative weight perturbation. The initial random guess procedure and the subsequent movements in random direction are repeated some predetermined number of times, which is specified by the user. The idea behind this method is similar to multivariate version of CART (i.e. coordinate descent over weights) but it picks the best of several random restarts.

\item \textbf{GUIDE:} Generalized, Unbiased, Interaction Detection and Estimation (GUIDE)  is the latest of the algorithms developed by Wei-Yin  \cite{Loh02a, Loh09a}, which improves upon his previous algorithms \citep{Loh14a}. GUIDE shares a similar tree inducing strategy as CART: greedily growing a large tree, and pruning it by cross validation. The major difference lies in the way features are selected and the choice of the split point. GUIDE utilizes various statistical tests in a single feature selection (axis-aligned) during node splitting with the aim of eliminating variable selection bias. GUIDE for classification also provides an option for bivariate linear splits in the nodes using only two features at a time, which are selected based on the chi-squared statistic of two variables.

\item \textbf{OCT:} \cite{BertsimDunn17a} have recently proposed to formulate a tree induction algorithm as an mixed-integer optimization that finds globally optimum tree and can be solved by using MIO solvers. However, this method has an exponential time complexity in the worst-case and thus can be applied only to small trees (usually up to depth 4). Otherwise, some early stopping criterion should be used which does not guarantee any optimality. We refer to this method as Optimal Classification Trees (OCT).

\item \textbf{CO2:} \cite{Norouz_15a} formulate a convex-concave upper bound on
the tree's empirical loss and optimize that loss using stochastic gradient descent. The initial tree structure must be provided and usually initialized using greedy procedure (similar to CART). The use of SGD enables efficient optimization for large scale datasets.

\item \textbf{TAO:} The TAO algorithm proposed in \cite{CarreirTavall18a} optimizes a decision tree with predetermined structure and can be trained to minimize the desired objective function such as misclassification error. Each iteration of TAO is guaranteed to decrease or leave unchanged the objective function. The algorithm can be applied to both axis aligned and oblique decision trees. Moreover, the algorithm can handle various penalty terms on objective function such as $\ell_1$-regularization which we briefly describe here (see \cite{CarreirTavall18a} for details). TAO assumes a given tree structure with initial parameter values (possibly random), and minimizes the following objective function jointly over the parameters $\bTheta = \{\btheta_i\}$ of all nodes $i$ of the tree:
\begin{equation}
  \label{e:objfcn}
  E(\bTheta) = \sum^N_{n=1}{ L(y_n,T(\x_n;\bTheta)) } + \lambda \sum_{\text{nodes $i$}}{ \norm{\w_i}_1 }
\end{equation}
where $\{(\x_n,y_n)\}^N_{n=1} \subset \bbR^D \times \{1,\dots,K\}$ is a training set of $D$-dimensional real-valued instances and their labels (in $K$ classes), $L(\cdot,\cdot)$ is the loss function (e.g. cross-entropy, MSE, 0/1 loss, etc.) and $T(\x;\bTheta)\mathpunct{:}\ \bbR^D \to \{1,\dots,K\}$ is the predictive function of the tree and $\btheta_i$ is parameters at a node $i$. For example, in case of oblique decision nodes, $\btheta_i$ is a hyperplane with weight vector $\w_i \in \bbR^D$ and bias $b_i \in \bbR$, which thus sends an input instance \x\ down its right child if $\w^T_i \x \ge b_i$ and down its left child otherwise.

The basis of the TAO algorithm is given by the separability condition theorem. It states that for any nodes $i$ and $j$ (internal or leaves) that are not descendants of each other (e.g.\ all nodes at the same depth) the error $E(\bTheta)$ in eq.~\eqref{e:objfcn} separates over $\btheta_i$ and $\btheta_j$. Since the loss function now separates algorithm can optimize eq.~\eqref{e:objfcn} over each node separately. This much simpler problem is referred as a ``reduced problem''. TAO algorithm applies alternating optimization over separable subsets of nodes: 

\begin{itemize}
	\item Optimizing over internal nodes is equivalent to optimizing a weighted binary classifier over $\btheta_i$ over the training instances $\{(\x_n,\overline{y}_n)\}$ that currently reach node $i$. Each such instance $\x_n$ is assigned a weight and pseudo label $\overline{y}_n \in \{\mathtt{-1},\mathtt{+1}\}$ based on the child whose subtree gives the better prediction for $\x_n$. Specifically, we send $\x_n$ to the left and right subtrees. All parameters in those subtrees are fixed and depending on which one gives correct (or better) output we assign a pseudo label (either $\mathtt{-1}$ or $\mathtt{+1}$). This pseudo label indicates where to send the given instance (either $\mathtt{left}$ or $\mathtt{right}$). We also assign a weight per sample because the loss of the best child is different for each instance.
	\item Optimizing over a leaf which is a $K$-class classifier on the training points that reach that particular leaf. In this paper, we focus on constant leaves. Therefore, the solution, in this case, will be the majority label of the training points that reach leaf $i$.
\end{itemize}

Depending on the decision node type, the first step can be solved either exactly (for axis-aligned nodes the best split is determined by enumerating over all the features) or approximated by surrogate loss (for oblique nodes it solves the linear binary classification problem).

\end{itemize}

\section{Experimental setup}
\label{s:expset}

\subsection{Algorithm-specific}

Below we describe each algorithm-specific experimental setup:

\begin{itemize}
\item \textbf{CART-R:} We use R implementation of CART called \texttt{rpart} ~\citep{Therneau_19a}. For each dataset during training we let the tree grow up to the max allowed depth of $30$ (max-depth constraint by  \texttt{rpart} ). For this, we set the ``minsplit" parameter to 1 and the complexity parameter (``cp'') to 0. Once the tree is fully grown we use \texttt{rpart} internal k-fold cross-validation (k=10), to get list of pruning parameters and choose best pruning parameter based on SE-1 rule (as suggested by \texttt{rpart} documentation). We report tests and train accuracy of the pruned tree.

\item \textbf{CART-P:} We also use the Python implementation of CART provided by scikit-learn \citep[version 0.22.2]{Pedreg_11a}. For each dataset, during training we let the tree grow full i.e. training error is zero. For this, we set the ``minsplit'' parameter to 1 and the complexity parameter (``ccp\_alpha'') to
0. Next, we find the best pruning parameter using k-fold cross-validation.

\item \textbf{C5.0:} We use the single-threaded Linux version of the C5.0 (provided by authors\footnote{\url{https://rulequest.com/download.html}}) written in C language. For each of the datasets, we apply a grid search on the k-fold validation set to get the best parameters. Specifically, we tune ``-c CF'' which controls the pruning severity and ``-m cases'' which is the minimum number of points to perform a node split. We use the default options for all other parameters. It worth to mention that empirically we have found that in many cases the tuned parameters are not far away from the default settings.

\item \textbf{GUIDE:} We use the GUIDE version 32.3 provided by the authors in the form of executables\footnote{\url{http://pages.stat.wisc.edu/~loh/guide.html}}. To run the executable, one needs to provide an input file, with all the parameters listed in a specific order. For \emph{axis-aligned classification}, we specify option (1) which gives univariate splits a higher priority, while for \emph{oblique trees}, we choose option (0) where linear splits are given a higher priority. GUIDE for regression uses only axis-aligned splits. We do not experiment with the kernel and nearest neighbor node models for classification and variants of linear regression used in regression trees, because they are beyond the scope of the conventional trees considered in this paper. The following hyperparameters were searched during cross validation: estimated or equal priors for class distribution, mean or median based cross validation, standard error for pruning, maximum number of splits and minimum node size. The latter two were the most important ones affecting the tree structure and accuracy, particularly for large datasets.
	
\item \textbf{OC1:} We use the implementation provided by the authors\footnote{\url{http://ccb.jhu.edu/software/oc1/oc1.tar.gz}} written in C. The default impurity measure used by their code is twoing criterion, and we only experiment with that option. For other impurity measures, one needs to change the code and recompile it again. We choose the option where only oblique hyperplanes are considered. During cross validation, we experiment with the number of random restarts and with the number of random jumps. For the most part, these two parameters affect the performance, but the results were not consistent. Moreover, this implementation only supports classification and thus we run OC1 on classification benchmarks.

\item \textbf{TAO:} We implement both axis-aligned and oblique versions of the TAO.
\begin{itemize}
  \item \emph{Oblique}: We use oblique (i.e. linear splits) decision trees with constant leaves. We take as an initial tree a deep enough, a complete binary tree with random parameters at each node. We use the fixed number of TAO iterations which is equal to $30$, and algorithm proceeds until the maximum number of iterations are reached (i.e. there is no other stopping criterion). We also use a simple grid search on  k-fold validation set to find the best hyperparameters. Specifically, we tune the ``$\lambda$'' parameter which controls sparsity of the tree and maximum depth of the initial tree. TAO algorithm is implemented in Python (version 3.5) without parallel processing in a single CPU. TAO uses an $\ell_1$-regularized logistic regression to solve the decision node optimization (using LIBLINEAR \citep{Fan_08a}) where the mentioned ``$\lambda$'' parameter is used as an regularization parameter ($C=1/\lambda$).
  \item \emph{Axis-aligned}: For the TAO axis-aligned trees, we use a decision tree with constant leaves (both for regression and classification). We initialized the tree with a pre-trained and pruned CART tree. Since TAO is implemented in Python 3.5, so we use scikit-learn implementation of CART as the initialization. We use a fixed number of TAO iterations which is equal to 30, and algorithm proceeds until the maximum number of iterations are reached or there is no more training error improvement up to a threshold of 1e-5. Unlike other algorithms, there are no hyper-parameters, as the tree is initialized with a pre-trained CART tree.
\end{itemize}

\item \textbf{OCT} We report the results from the corresponding paper (for both axis aligned and oblique) since the implementation is not available online.

\item \textbf{CO2} We report the results from the corresponding paper since the implementation is not available online.

\end{itemize}

We ran all experiments on a single Linux PC with the following specifications: OS - Ubuntu 18.04 LTS, CPU - 8 $\times$ Intel Core i7-7700 3.60GHz, Memory - 16 GiB DDR4 3600 MHz.

\subsection{Datasets}
\label{s:datasets}

\begin{table}
  \centering
  \begin{tabular}{@{}l@{\hspace{0.5ex}}|lrrrcl@{}}
    \toprule    
    & Dataset & $N_{\text{train}}$ & $N_{\text{test}}$ & $D$ & $K$ & Comments\\
    \midrule
    & Iris&	120&	30&	4&	3 & ---\\
    & Wine&	142&	36&	13&	3 & ---\\
    & Dermatology&	293&	73&	34&	6 & ---\\    
    & Balance scale&	500&	125&	4&	3 & ---\\
    & Breast Cancer&	559&	140&	9&	2 & ---\\    
    & Blood Trans&	598&	150&	4&	2 & ---\\
    & German&	800&	200&	20&	2 & categorical features encoded as one-hot\\
    & Banknote auth&	1098&	274&	4&	2 & ---\\
    & Contraceptive&	1178&	295&	9&	3 & categorical features encoded as one-hot\\    
    & Car Eval&	1382&	346&	6&	4 & categorical features encoded as one-hot\\
    \raisebox{11pt}[0pt][0pt]{\rotatebox{90}{\makebox[0pt][c]{classification}}}
    & Segment&	1848&	462&	19&	7 & ---\\
    & Spambase&	3681&	920&	57&	2 & ---\\
    & Optical recog&	3823&	1797&	64&	10 & ---\\
    & Landsat&	4435&	2000&	36&	6 & ---\\
    & Pendigits&	7494&	3498&	16&	10 & ---\\
    & Letter$^{\text{a}}$&	16000&	4000&	16&	26 & ---\\
    & Connect4&	54046&	13511&	126&	3 & ---\\
    & MNIST (pixels)$^{\text{a}}$&	60000&	10000&	784&	10 & grayscale image with pixels in [0,1]\\
    & MNIST (LeNet5)$^{\text{b}}$&	60000&	10000&	800&	10 & output of LeNet5-conv2 on
    MNIST pixels \\
    & SensIT$^{\text{a}}$&	78823&	19705&	100&	3 & ---\\
    \midrule
    & concrete & 687 & 343 & 8 & 1 & ---\\
    & airfoil & 1002 & 501 & 5 & 1 & ---\\
    & abalone & 2506 & 1671 & 8 & 1 & categorical features encoded as one-hot\\
    & cpuact$^{\text{c}}$ & 4915 & 3277 & 21 & 1 & ---\\
    \raisebox{11pt}[0pt][0pt]{\rotatebox{90}{\makebox[0pt][c]{regression}}}
    & ailerons$^{\text{d}}$ & 7154 & 6596 & 40 & 1 & ---\\
    & CT slice & 42800 & 10700 & 384 & 1 & ---\\
    & YearPredictionMSD & 463715 & 51630 & 90 & 1 & ---\\
    \bottomrule
    \multicolumn{7}{l}{$^{\text{a}}$\scriptsize\url{https://www.csie.ntu.edu.tw/~cjlin/libsvmtools/datasets/multiclass.html}}\\
    \multicolumn{7}{l}{$^{\text{b}}$\scriptsize\url{https://faculty.ucmerced.edu/mcarreira-perpinan/teaching/CSE176/Labs/datasets/}}\\
    \multicolumn{7}{l}{$^{\text{c}}$\scriptsize\url{http://www.cs.toronto.edu/~delve/data/comp-activ/desc.html}}\\
    \multicolumn{7}{l}{$^{\text{d}}$\scriptsize\url{https://www.dcc.fc.up.pt/~ltorgo/Regression/DataSets.html}}
  \end{tabular}
  \caption{Specs of the datasets used in our experiments. $N$ is sample size, $D$ is feature dimensionality and $K$ is output dimension for regression and number of classes for classification. 
All datasets are from the UCI collection \citep{Lichman13a} unless otherwise indicated by a footnote.}
  \label{t:data-descr}
\end{table}

Table~\ref{t:data-descr} summarizes all datasets used in this study with their corresponding sources. All datasets are available in the public domain. The exception is ``MNIST (LeNet5)''. This dataset consists of features extracted by ``conv2'' layer of a pre-trained  LeNet5 \citep{Lecun_98a}  neural network for all MNIST images. Similar to MNIST, it also has $60000$ training and $10000$ test instances. Each instance has $800$ non-negative real-valued attributes. For some of the datasets from UCI which do not have separate test set, we shuffle the entire dataset and keep 20\% of the entire data as the test set. We repeat the training procedure $10$ times for each dataset, reshuffling the training data each time.

\section{Results}
\label{s:result}
\subsection{Classification}

We summarize the results for classification datasets in Table~\ref{t:comp-axis-test}-\ref{t:comp-oblique-test}. We report the train/test average accuracy (in \%) and stdev over 10 repeats. Although we report the training error for reference, \emph{the important error to consider is the test error}. Because the training error can be trivially made zero with a tree by simply growing it large enough that each leaf contains instances of the same class, at the cost of overfitting. Please note that some results in the tables are missing, because they were not provided in the original papers.

First, consider the case of an axis-aligned split. OCT shows a good performance only on relatively small datasets like Iris or Wine. We think that this is due to limited depth of the tree. Solving for deeper trees require a huge amount of time since OCT solves tree optimization problem exactly which is NP-hard, unless some early stopping criteria applied which does not guarantee an optimal solution. Therefore, we observe that OCT shows the worst performance in most cases. Next, both versions of the CART and C5.0 perform similarly in terms of test accuracy with a little favor to C5.0. GUIDE also performs similar to CART and C5.0. These methods share the same flavor of greedy recursive partitioning of the input space and thus it is expected that they have similar performance. Finally, TAO outperforms all algorithms in most datasets. In the very few datasets where TAO is not the winner, the difference with the winner is very small, and is due to a handful of instances, since these are small datasets in sample size (and dimension).

Second, consider the case of oblique decision trees. Here, we can observe similar behavior as with axis-aligned trees. TAO outperforms all other methods (including non-greedy approaches like CO2) in most dataset and often by considerable margin. The accuracy margin between TAO and the other methods becomes more as the dataset complexity grows like in MNIST. For instance, in the case of ``MNIST (pixels)'' and ``MNIST (LeNet5)'' not only the dataset size is big ($60000$ training data points) but also the number of attributes is very high. In the few cases where another algorithm produced a better tree, we could always further improve TAO by the following way: \emph{TAO can take any given initial tree and improve its training loss or leave it unchanged} (unlike traditional top-down induction algorithms such as CART or C5.0, which build the tree from scratch). However, we do not use this advantage, and in all our experiments we initialize TAO oblique trees randomly.

Since decision trees are considered as interpretable models, it is important to also compare the size of the trained trees since if a decision tree is deep and has a large number of nodes then it becomes extremely difficult to interpret it. Moreover, the larger tree has large inference time and need more space. For this, in Table~\ref{t:comp-all-depth}-\ref{t:comp-all-leaves} we compare the average maximum depth and average number of leaves for both axis-aligned and oblique trees. Results makes sense: bigger datasets require larger trees (more depth and more leaves) and oblique trees generate more compact trees but with more complex node structure (i.e. generic hyperplane instead of axis-aligned split). The exception is OCT since authors have limited the maximum depth due to reasons described above. C5.0 usually generates larger trees compared to CART given that both perform similarly in terms of test accuracy. In general, decision trees obtained from TAO have comparable or smaller sizes as from the other methods. In some datasets (like Letter), the number of leaves and maximum depth are quite large for TAO axis aligned but it is due to large number of classes. More compact models can be obtained by using oblique tree. 

\begin{table}[p]
  \centering
  \begin{tabular}{@{}l@{}l|r@{$\pm$}rr@{$\pm$}rr@{$\pm$}rr@{$\pm$}rr@{$\pm$}rr@{}}
  \toprule
  Dataset & & \multicolumn{2}{c}{TAO}&	\multicolumn{2}{c}{CART-R}&	\multicolumn{2}{c}{CART-P}&	\multicolumn{2}{c}{C5.0}&	\multicolumn{2}{c}{GUIDE}&	OCT\\
  \midrule
  \raisebox{-6pt}[0pt][0pt]{Iris} &train &	97.03&0.65  &	97.08&1.19  &	97.02&0.99  &	99.03&1.25&	97.00 &1.35 &---\\
  & test &	 \textcolor{blue}{95.41} &3.81 &	 93.33&3.15 &	 \textcolor{DarkRed}{94.75}&3.20 &	92.64&8.00&	93.67 & 4.58&	93.5\\
  \raisebox{-6pt}[0pt][0pt]{Wine} &train &	96.01&1.49  &	96.97&2.63  &	95.86&1.22  &	96.68&1.58&	96.76 &2.28&---\\  
  & test &	 91.21&3.64 &	 90.00&3.97 &	 90.10&2.21 &	85.19&15.95& \textcolor{DarkRed}{91.94} & 5.04&	\textcolor{blue}{94.2}\\
  \raisebox{-6pt}[0pt][0pt]{Dermatology} &train &	98.61&0.90  &	97.19&1.02  &	98.54&0.88  &	97.14&0.72&	97.33 &0.68&---\\
  & test &	\textcolor{blue}{96.14}&2.72 &	 93.51&1.66 &	 94.44&3.41 &	90.44&4.78&	\textcolor{DarkRed}{95.54} & 3.08&	89.2\\
  \raisebox{-6pt}[0pt][0pt]{Balance scale} &train &	85.95&1.28 &	85.94&0.42&	82.66&0.92 &	88.38&1.43&	86.42 &2.81&---\\
  & test &	\textcolor{blue}{82.21} &3.36 &	78.96&0.34&	 \textcolor{DarkRed}{79.62}&2.09 &	78.19&1.43&	77.44 & 3.26&	71.6\\
  \raisebox{-6pt}[0pt][0pt]{Breast Cancer} &train &	95.39&2.44 &	96.10&0.01&	95.26&1.22 &	97.36&0.61&	96.62 &0.79&---\\
  & test &	\textcolor{blue}{95.91}&1.54 &	94.57&0.02&	 93.64&2.80 &	\textcolor{DarkRed}{94.83}&0.90&	94.57 & 1.75&	91.5\\
  \raisebox{-6pt}[0pt][0pt]{Blood Trans} &train &	82.11&1.47 &	76.45&0.01&	79.42&2.31 &	79.69&0.59&	80.90 &1.97&---\\
  & test &	 77.89&3.27 &	75.20&0.02&	 76.45&2.77 &	\textcolor{DarkRed}{78.40}&2.45&	\textcolor{blue}{78.80} & 3.26&	77.0\\
  \raisebox{-6pt}[0pt][0pt]{Banknote auth} &train &	99.67&0.19 &	99.45&0.02&	99.30&0.24 &	99.63&0.12&	99.38 &0.08&---\\
  & test &	 97.98&0.91 &	97.93&0.06&	 96.33&2.19 &	\textcolor{blue}{98.70}&0.68&	\textcolor{DarkRed}{98.25} & 1.69&	90.7\\
  \raisebox{-6pt}[0pt][0pt]{Contraceptive} &train & 54.45&1.38 &	57.53&1.05 & 53.88&1.27 &	76.15&1.41&	58.79 &1.89&---\\  
  & test  & \textcolor{blue}{56.03} &2.89 & 54.37&1.83 & 54.25 &4.43 &	49.39&4.79&	\textcolor{DarkRed}{55.19} & 2.13&	53.3\\  
  \raisebox{-6pt}[0pt][0pt]{Car Eval} &train &	99.94&0.12  &	98.36&1.41  &	99.91&0.12  &	92.71&3.85&	69.97 &0.48&---\\  
  & test &	\textcolor{blue}{96.67}&2.67 &	  96.35&1.90 &	\textcolor{DarkRed}{96.51}&2.69 &	87.59&4.19&	70.23 & 1.93&	78.8\\
  \raisebox{-6pt}[0pt][0pt]{Segment} &train &	98.39&0.54 &	98.87&0.69 &	98.34&0.58 &	98.42&0.88&	98.31 &0.44&---\\
  & test &	 \textcolor{blue}{96.18}&1.41 &	\textcolor{DarkRed}{95.91}&0.11 &	 92.17&3.85 &	94.86&1.69&	95.37 & 1.11&	---\\
  \raisebox{-6pt}[0pt][0pt]{Spambase} &train &	94.27&1.52 &	94.96&0.01&	93.18&2.71 &	96.18&0.38&	95.36 &0.85&---\\
  & test &	 \textcolor{DarkRed}{92.65}&1.18 &	91.92&0.01&	 89.62&3.28 &	\textcolor{blue}{92.85}&0.84&	92.77 & 0.64&	86.1\\
  \raisebox{-6pt}[0pt][0pt]{Optical recog} &train &	98.48&0.06  &	95.33&2.18  &	97.85&0.12  &	96.76&0.84&	94.64 &1.10&---\\
  & test &	\textcolor{blue}{86.08}&0.22 &	 84.26&1.03 &	 \textcolor{DarkRed}{85.19}&0.11 &	80.33&5.32&	84.56 & 1.32&	54.7\\
  \raisebox{-6pt}[0pt][0pt]{Landsat} &train &	96.97&0.26  &	91.13&0.49  &	97.37&0.12  &	95.19&0.71&	90.05 &0.86&---\\
  & test &	\textcolor{blue}{86.10}&0.25 & \textcolor{DarkRed}{85.94}&0.27 &	  85.21&0.23 &	84.12&1.09&	85.24 & 0.58&	78.0\\
  \raisebox{-6pt}[0pt][0pt]{Pendigits} &train &	98.87&0.26 &	99.09&0.29 &	98.79&0.13 &	98.98&0.09&	97.90 &0.50&---\\
  & test &	\textcolor{blue}{92.52}&0.24 &	\textcolor{DarkRed}{91.62}&0.55 &	 90.19&0.38 &	91.32&0.71&	89.93 & 0.77&	---\\
  \raisebox{-6pt}[0pt][0pt]{Letter} &train &	96.38&0.03 &	94.30&0.01&	95.93&0.11 &	97.97&0.14&	93.54 &0.59&---\\
  & test &	\textcolor{blue}{86.39}&0.12 &	86.04&0.04&	 \textcolor{DarkRed}{86.07}&0.14 &	85.26&0.33&	83.93 & 0.48&	---\\
  \raisebox{-6pt}[0pt][0pt]{Connect4} &train &	86.99&1.74  &	82.94&1.08  &	83.63&0.13  &	82.60&0.93&	71.76 &0.24&---\\
  & test &	\textcolor{blue}{79.88} &1.53 &	\textcolor{DarkRed}{78.29}&0.21 &	  77.55&1.18 &	77.84&1.41&	71.66 & 0.32&	---\\ 
  \raisebox{-6pt}[0pt][0pt]{MNIST (pixels)} &train &	93.53&0.24 &	92.54&0.03&	93.12&0.15 &	94.52&0.23&	75.76 &0.40&---\\
  & test &	\textcolor{blue}{88.52} &0.19 &	88.03&0.07&	 88.05&0.02 &	\textcolor{DarkRed}{88.31}&0.35&	78.52 &0.20&	---\\
  \raisebox{-6pt}[0pt][0pt]{MNIST (LeNet5)} &train &	96.94&0.05 &	95.71&0.04&	96.64&0.08 &	97.89&0.14&	84.15 &0.28&---\\
  & test &	\textcolor{blue}{93.52}&0.03 &	93.31&0.05&	 93.32&0.07 &	\textcolor{DarkRed}{93.48}&0.21&	85.25 & 0.28&	---\\
  \raisebox{-6pt}[0pt][0pt]{SensIT} &train &	83.56&0.12 &	84.38&0.01&	82.82&0.23 &	86.66&0.11&	79.05 &0.23&---\\  
  & test &	\textcolor{blue}{81.84}&1.11 & \textcolor{DarkRed}{81.71} &0.01&	 81.00&0.19 &	81.41&0.04&	78.52 & 0.20&	---\\
  \midrule
  wins & train & \multicolumn{2}{c}{6} & \multicolumn{2}{c}{3} & \multicolumn{2}{c}{1} & \multicolumn{2}{c}{9} & \multicolumn{2}{c}{1} &  0 \\
  \textbf{wins} & \textbf{test} & \multicolumn{2}{c}{15} & \multicolumn{2}{c}{0} & \multicolumn{2}{c}{0} & \multicolumn{2}{c}{2} & \multicolumn{2}{c}{1} &  1\\
  \bottomrule
  \end{tabular}
  \caption{Train and test accuracy (\%, avgstdev over 10 repeats) for decision tree learning algorithms with axis-aligned splits. Method names are in section~\ref{s:expset}. Colors within each test row represents: Blue - the best performing method, Dark Red - the 2nd best performing method. Authors in OCT paper do not report training error or errorbars, only the average test error. }
  \label{t:comp-axis-test}
\end{table}

\begin{table}[p]
  \centering
  \begin{tabular}{@{}l@{}l|r@{$\pm$}rr@{$\pm$}rr@{$\pm$}rrr@{}}
  \toprule
  Dataset& & \multicolumn{2}{c}{TAO}&	\multicolumn{2}{c}{OC1}&	\multicolumn{2}{c}{GUIDE}&	\multicolumn{1}{c}{OCT}&	\multicolumn{1}{c}{CO2}\\
  \midrule
  \raisebox{-6pt}[0pt][0pt]{Iris} &train &	96.89&9.05&	85.42 &15.84&	98.42 &0.58&	---&---\\
  & test &	\textcolor{DarkRed}{94.40}&5.12&	85.67 &14.53&	94.33 &3.00&	\textcolor{blue}{95.1}&	---\\
  \raisebox{-6pt}[0pt][0pt]{Wine} &train &	98.22&5.33&	89.30 &10.25&	97.75 &1.50&	---&---\\
  & test &	\textcolor{DarkRed}{92.00}&9.38&	84.45 &8.89&	\textcolor{blue}{93.33} &5.00&	91.6&	---\\
  \raisebox{-6pt}[0pt][0pt]{Dermatology} &train &	95.10&3.96&	91.58 &6.10&	98.60 &0.50&	---&---\\
  & test &	92.27&8.57&	84.46 &7.79&	\textcolor{blue}{97.84} &1.73&	\textcolor{DarkRed}{92.6}&	---\\  
  \raisebox{-6pt}[0pt][0pt]{Balance scale} &train &	91.68&0.72&	93.22 &2.17&	91.72 &3.94&	---&---\\
  & test &	\textcolor{DarkRed}{88.48}&2.56&	\textcolor{blue}{88.96} &2.29&	85.60 &5.68&	87.6&	---\\
  \raisebox{-6pt}[0pt][0pt]{Breast Cancer} &train &	98.21&0.79&	82.99 &12.16&	96.82 &0.51&	---&---\\
  & test &	\textcolor{blue}{97.71}&1.04&	81.07 &12.75&	\textcolor{DarkRed}{95.64} &1.61&	94.0&	---\\
  \raisebox{-6pt}[0pt][0pt]{Blood Trans} &train &	81.74&0.89&	80.33 &2.69&	81.02 &0.63&	---&---\\
  & test &	\textcolor{DarkRed}{78.93}&3.12&	77.93 &3.72&	\textcolor{blue}{79.87} &3.33&	77.4&	---\\
  \raisebox{-6pt}[0pt][0pt]{German} &train &	82.90&0.71&	78.44 &5.85&	70.18 & 1.07 & ---&---\\
  & test &	\textcolor{blue}{81.24}&0.87&	68.65 &4.17&	70.15& 2.17 &	\textcolor{DarkRed}{71.0}&	---\\  
  \raisebox{-6pt}[0pt][0pt]{Banknote auth} &train &	99.83&0.33&	94.12 &12.91&	99.63 &0.27&	---&---\\
  & test &	\textcolor{blue}{99.18}&0.14&	91.64 &13.57&	\textcolor{DarkRed}{98.80} &0.59&	98.7&	---\\
  \raisebox{-6pt}[0pt][0pt]{Contraceptive} &train &	66.04&7.24&	61.44 &5.71&	57.58 &0.93&	---&---\\
  & test &	\textcolor{blue}{57.47}&3.18&	49.66 &3.06&	\textcolor{DarkRed}{56.58} &2.58&	53.3&	---\\
  \raisebox{-6pt}[0pt][0pt]{Car Eval} &train &	94.27&4.12&	97.35 &2.92&	69.97 &0.48&	---&---\\
  & test &	\textcolor{DarkRed}{91.55}&4.66&	\textcolor{blue}{95.49} &2.32&	70.23 &1.93&	87.5&	---\\
  \raisebox{-6pt}[0pt][0pt]{Segment} &train &	99.48&0.21&	91.61 &8.84&	98.41 &0.41&	---& 97\\
  & test &	\textcolor{blue}{96.48}&1.31&	88.53 &7.47&	95.48 &1.02&	---&	\textcolor{DarkRed}{96}\\
  \raisebox{-6pt}[0pt][0pt]{Spambase} &train &	95.55&0.47&	80.72 &16.51&	95.74 &0.99&	---&---\\
  & test &	\textcolor{blue}{93.31}&1.22&	78.20 &15.48&	\textcolor{DarkRed}{92.24} &0.59&	86.6&	---\\
  \raisebox{-6pt}[0pt][0pt]{Optical recog} &train &	97.68&0.59&	72.50 &19.62&	94.54 &1.36&	---&---\\
  & test &	\textcolor{blue}{91.27}&1.74&	62.00 &17.60&	\textcolor{DarkRed}{79.19} &1.20&	54.3&	---\\
  \raisebox{-6pt}[0pt][0pt]{Landsat} &train &	94.45&0.49&	80.25 &2.20&	91.54 &1.29&	---&---\\
  & test &	\textcolor{blue}{87.81}&0.88&	73.54 &2.00&	\textcolor{DarkRed}{85.97} &0.80&	78.2&	---\\  
  \raisebox{-6pt}[0pt][0pt]{Pendigits} &train &	99.81&0.13&	91.72 &7.81&	98.85 &0.14&	---& 96\\
  & test &	\textcolor{blue}{96.80}&0.70&	84.42 &7.02&	91.80 &0.69&	---&	\textcolor{DarkRed}{92}\\
  \raisebox{-6pt}[0pt][0pt]{Connect4} &train &	82.40&0.53&	79.02 &1.45&	72.11 &0.31&	---& 81\\
  & test &	\textcolor{blue}{81.09}&0.39&	75.42 &0.64&	72.01 &0.36&	---&	\textcolor{DarkRed}{78}\\
  \raisebox{-6pt}[0pt][0pt]{Letter} &train &	95.39&0.24&	75.85 &3.80&	90.80 &1.05&	---& 94\\
  & test &	\textcolor{blue}{89.15}&0.88&	65.81&4.83&	82.65 &0.90&	---&	\textcolor{DarkRed}{87}\\
  \raisebox{-6pt}[0pt][0pt]{MNIST (pixels)} &train &	98.43&0.07&	78.62 &9.62&	73.02&0.79&	---& 94\\
  & test &	\textcolor{blue}{94.74}&0.11&	74.34 &9.94&	73.79 &0.91&	---&	\textcolor{DarkRed}{90}\\
  \raisebox{-6pt}[0pt][0pt]{MNIST (LeNet5)} &train &	99.98&0.01&	89.52 &14.76& 84.15&0.28&	---&---\\
  & test &	\textcolor{blue}{98.22}&0.18&	\textcolor{DarkRed}{87.97} &14.24&	85.25&0.28&	---&	---\\
  \raisebox{-6pt}[0pt][0pt]{SensIT} &train &	85.68&0.13&	76.10 &12.69&	79.64 &0.28 & --- & 83\\
  & test &	\textcolor{blue}{85.12}&0.20&	73.70 &11.31&	79.25&0.33&	---&	\textcolor{DarkRed}{82}\\
  \midrule
  wins & train & \multicolumn{2}{c}{15} & \multicolumn{2}{c}{2} & \multicolumn{2}{c}{3} & 0 & 0 \\
  \textbf{wins} & \textbf{test}& \multicolumn{2}{c}{14} & \multicolumn{2}{c}{2} & \multicolumn{2}{c}{3} & 1 & 0 \\
  \bottomrule
  \end{tabular}
  \caption{Similar to \ref{t:comp-axis-test} but for decision trees with oblique splits.}
  \label{t:comp-oblique-test}
\end{table}

\begin{table}
  \centering
  \begin{tabular}{@{}l|r@{\hspace{1.0ex}}r@{\hspace{1.0ex}}r@{\hspace{1.5ex}}r@{\hspace{1.2ex}}r@{\hspace{1.0ex}}r@{\hspace{1.0ex}}|r@{\hspace{1.0ex}}r@{\hspace{1.0ex}}r@{\hspace{1.0ex}}r@{\hspace{1.0ex}}r@{}}
  \toprule
  & \multicolumn{6}{c|}{axis-aligned} & \multicolumn{5}{c}{oblique} \\
  \midrule
  Dataset&	TAO&	CART-P&	CART-R&	C5.0&	GUIDE&	OCT&	TAO&	OC1&	GUIDE&	OCT&	CO2\\
  \midrule
  Iris&	3.0&	3.0&	2.5&	1.5&	2.3&	4.0&	3.0&	1.6&	2.3&	4.0&	---\\
  Wine&	2.8&	2.8&	3.0&	1.2&	2.7&	4.0&	5.0&	2.0&	2.4&	4.0&	---\\
  Dermatology&	7.0&	7.0&	6.1&	4.7&	6.7&	4.0&	7.0&	3.0&	5.1&	4.0&	---\\  
  Balance scale&	7.2&	7.2&	6.7&	7.1&	6.3&	4.0&	3.0&	2.9&	5.8&	4.0&	---\\
  Breast Cancer&	3.4&	3.4&	3.2&	4.0&	4.0&	4.0&	3.0&	2.5&	1.8&	4.0&	---\\
  Blood Trans&	7.4&	7.4&	0.0&	2.5&	4.8&	4.0&	5.0&	3.9&	2.4&	4.0&	---\\  
  Banknote auth&	6.0&	6.0&	5.8&	5.8&	6.7&	4.0&	3.0&	4.2&	5.1&	4.0&	---\\
  Contraceptive & 4.6 & 4.6	&	4.3&	11.1&	5.6&	4.0&	5.0&	5.1&	4.0&	4.0&	---\\
  Car Eval&	12.7&	12.2&	11.8&	3.6&	4.6&	4.0&	4.0&	3.5&	3.6&	4.0&	---\\
  Segment&	14.0&	14.0&	13.8&	7.3&	13.2&	---&	8.0&	5.3&	15.2&	---&	8.0\\      
  Spambase&	14.4&	14.4&	10.7&	14.7&	12.3&	4.0&	4.0&	3.56&	15.4&	4.0&	---\\
  Optical recog&	12.0&	12.0&	11.7&	10.8&	13.4&	4.0&	7.0&	4.2&	13.7&	4.0&	---\\
  Landsat&	12.0&	12.0&	11.8&	11.1&	9.9&	4.0&	7.0&	6.3&	17.0&	4.0&	---\\
  Pendigits&	15.2&	15.2&	13.8&	13.6&	14.0&	---&	8.0&	5.9&	20.1&	---&	12.0\\
  Letter&	27.0&	27.0&	26.0&	17.0&	23.4&---	&	11.0&	10.2&	28.0&	---&	12.0\\
  Connect4&	33.2&	33.2&	27.7&	16.8&	16.7&	---&	8.0&	8.6&	17.9&	---&	16.0\\
  MNIST (pixels)&	19.0&	19.0&	18.3&	19.0&	8.5&---	&	8.0&	5.0&	14.9&	---&	14.0\\
  MNIST (LeNet5)&	17.0&	17.0&	18.6&	15.2&	9.3&---	&	8.0&	4.4&	9.3&---	&	---\\
  SensIT&	12.0&	12.0&	14.0&	15.2&	9.8& ---	&	7.0&	8.1&	10.8&	---&	6.0\\
  \bottomrule
  \end{tabular}
  \caption{Average maximum depths over 10 repetitions for both axis-aligned and oblique classification trees.}
  \label{t:comp-all-depth}
\end{table}

\begin{table}
  \centering
  \begin{tabular}{@{}l|rrrrr|rrr@{}}
  \toprule
  & \multicolumn{5}{c|}{axis-aligned} & \multicolumn{3}{c}{oblique} \\
  \midrule
  Dataset&	TAO&	CART&	CART&	C5.0&	GUIDE&	TAO&	OC1&	GUIDE\\
  \midrule
  Iris&	4.4&	4.4&	3.6&	3.5&	3.3&	8.1&	2.6&	3.3\\
  Wine&	5.6&	5.8&	5.5&	3.5&	4.6&	16.0&	3.2&	3.5\\
  Dermatology&	9.6&	9.6&	7.2&	6.7&	8.6&	64.0&	5.0&	6.2\\
  Balance scale&	24.6&	24.6&	22.6&	27.8&	18.1&	5.6&	4.3&	8.3\\
  Breast Cancer&	5.4&	5.6&	5.5&	9.0&	7.6&	7.8&	4.0&	3.2\\
  Blood Trans&	20.0&	20.8&	1.0&	4.6&	8.2&	10.8&	5.7&	3.5\\      
  Banknote auth&	19.2&	20.2&	14.0&	14.3&	19.0&	7.4&	9.3&	8.3\\
  Contraceptive & 7.2 & 7.4 &	7.6&	89.6&	12.7&	24.4&	9.3&	7.3\\
  Car Eval&	68.0&	68.0&	56.7&	41.2&	6.7&	14.3&	5.6&	5.0\\
  Segment&	38.2&	39.2&	41.3&	21.0&	39.7&	135.0&	11.0&	31.2\\  
  Spambase&	53.6&	55.0&	41.7&	68.6&	53.5&	14.8&	5.1&	48.0\\
  Optical recog&	193.8&	198.2&	107.2&	72.6&	126.6&	57.4&	9.6&	138.7\\
  Landsat&	231.4&	236.4&	57.6&	67.1&	50.8&	70.6&	13.4&	50.5\\
  Pendigits&	177.2&	183.6&	153.6&	129.0&	161.5&	146.0&	19.4&	135.5\\
  Letter&	1550.8&	1579.6&	920.8&	1343.0&	994.8&	1077.6&	88.7&	673.1\\
  Connect4&	5336.0&	5743.8&	1212.6&	813.0&	38.4&	210.0&	32.8&	26.5\\
  MNIST (pixels)&	899.5&	899.5&	805.4&	941.6&	58.8&	177.8&	12.8&	38.5\\
  MNIST (LeNet5)&	484.0&	484.0&	363.2&	582.0&	42.4&	166.8&	11.8&	42.4\\
  SensIT&	152.0&	152.0&	239.5&	410.0&	41.6&	69.2&	21.8&	24.3\\
  \bottomrule
  \end{tabular}
  \caption{Average number of leaves over 10 repetitions for both axis-aligned and oblique classification trees.}
  \label{t:comp-all-leaves}
\end{table}

\subsection{Regression}

We also perform comparison on regression datasets. Table~\ref{t:comp-regr-test} reports the results. All reported errors are rooted mean squared error (RMSE): $E = \sqrt{\frac{1}{NK}\sum^N_{n=1}{\norm{\y_n - \hat{\y}_n}^2}}$ unless otherwise specified, where $N$ is the sample size, $K$ is the output dimension, and \y\ and $\hat{\y}$ are the ground truth and predicted vectors, respectively. Please, note that GUIDE for regression uses only axis-aligned splits. Results clearly demonstrate that TAO trees outperform all the other methods. Especially, TAO oblique shows superior accuracy and often by a large margin followed by TAO axis aligned. Moreover, the resulting tree size (see Table~\ref{t:comp-regr-depth-leafno}) is small in TAO oblique which makes it easier to interpret. In the few cases where TAO oblique tree performs worse than the axis-aligned tree, we believe this is due to the dataset size. Furthermore, we could always further improve TAO by initializing from axis-aligned tree as was described in the previous section.

\begin{table}[h]
  \centering
  \begin{tabular}{@{}l@{}l|r@{\hspace{0.5ex}}l|r@{\hspace{0.5ex}}lr@{\hspace{0.5ex}}lr@{\hspace{0.5ex}}lr@{\hspace{0.5ex}}l@{}}
  \toprule
  & & \multicolumn{2}{c|}{oblique} & \multicolumn{8}{c}{axis-aligned} \\
  \midrule
  Dataset& & \multicolumn{2}{c|}{TAO}&	\multicolumn{2}{c}{TAO}&	\multicolumn{2}{c}{CART-R}&	\multicolumn{2}{c}{CART-P}&	\multicolumn{2}{c}{GUIDE}\\
  \midrule
  \raisebox{-6pt}[0pt][0pt]{concrete} &train &	3.91&$\pm$0.11&	3.06&$\pm$5.36&	3.93&$\pm$2.67&	3.07&$\pm$2.31&	5.46 &$\pm$ 0.37\\
  & test &	7.41&$\pm$0.12&	\textcolor{blue}{7.20} &$\pm$3.17 &	7.23 &$\pm$3.08 &	\textcolor{DarkRed}{7.22} &$\pm$3.13 &	7.50 &$\pm$ 0.27\\
  \raisebox{-6pt}[0pt][0pt]{airfoil} &train &	3.01&$\pm$0.29&	0.47&$\pm$0.10&	0.72&$\pm$0.12&	0.52&$\pm$0.10&	2.44 &$\pm$ 0.10\\
  & test &	3.13&$\pm$0.38&	\textcolor{blue}{2.73} &$\pm$0.62 &	2.77 &$\pm$0.86 &	\textcolor{DarkRed}{2.75} &$\pm$0.62 &	3.20 &$\pm$ 0.19\\
  \raisebox{-6pt}[0pt][0pt]{abalone} &train &	2.11&$\pm$0.02&	2.29&$\pm$0.12&	2.31&$\pm$0.25&	2.32&$\pm$0.11&	2.21 &$\pm$ 0.05\\
  & test &	\textcolor{blue}{2.18}&$\pm$0.05&	2.32 &$\pm$0.58 &	2.38 &$\pm$0.31 &	2.34 &$\pm$0.59 &	\textcolor{DarkRed}{2.30} &$\pm$ 0.09\\
  \raisebox{-6pt}[0pt][0pt]{cpuact} &train &	2.47&$\pm$0.07&	2.68&$\pm$0.69&	2.91&$\pm$0.71&	2.71&$\pm$0.65&	10.00 &$\pm$ 0.88\\
  & test &	\textcolor{blue}{2.71}&$\pm$0.04&	\textcolor{DarkRed}{3.26} &$\pm$0.51 &	3.36 &$\pm$1.32 &	3.28 &$\pm$0.44 &	10.99 &$\pm$ 1.54\\
  \raisebox{-6pt}[0pt][0pt]{ailerons} &train &	1.65&$\pm$0.02&	2.39&$\pm$0.00&	1.81&$\pm$0.12&	2.83&$\pm$0.23&	1.86 &$\pm$ 0.02\\
  & test &	\textcolor{blue}{1.76}&$\pm$0.02&	2.55 &$\pm$0.00 &	2.21 &$\pm$0.63 &	2.85 &$\pm$0.57 &	\textcolor{DarkRed}{2.06} &$\pm$ 0.02\\
  \raisebox{-6pt}[0pt][0pt]{CT slice} &train  &	1.42&$\pm$0.04&	1.01&$\pm$0.04&	1.12&$\pm$0.15&	1.06&$\pm$0.06&	8.12 &$\pm$ 0.17\\
  & test  &	\textcolor{blue}{1.54}&$\pm$0.05&	\textcolor{DarkRed}{2.66}&$\pm$0.04 &	2.91 &$\pm$0.95 &	2.69 &$\pm$0.03 &	8.23 &$\pm$ 0.20\\
  \raisebox{-6pt}[0pt][0pt]{YearPredictionMSD} &train   &	8.91&$\pm$0.03&	 9.71&$\pm$0.31&	9.73&$\pm$0.20&	9.71&$\pm$0.24&	9.78 &$\pm$ 0.01\\
  & test &	\textcolor{blue}{9.11}&$\pm$0.05&	 \textcolor{DarkRed}{9.76} &$\pm$0.11 &	9.81&$\pm$0.31 &	9.79 &$\pm$0.54 &	9.83 &$\pm$ 0.01\\
  \midrule
  wins & train & \multicolumn{2}{c|}{4} & \multicolumn{2}{c}{3} & \multicolumn{2}{c}{0} & \multicolumn{2}{c}{0} & \multicolumn{2}{c}{0} \\
  \textbf{wins} & \textbf{test} & \multicolumn{2}{c|}{5} & \multicolumn{2}{c}{2} & \multicolumn{2}{c}{0} & \multicolumn{2}{c}{0} & \multicolumn{2}{c}{0} \\
  \bottomrule
  \end{tabular}
  \caption{Train and test rooted mean squared error (RMSE) for different methods on regression datasets. Results for ailerons scaled to $\smash{E\times10^{-4}}$.}
  \label{t:comp-regr-test}
\end{table}

\begin{table}[h]
  \centering
  \begin{tabular}{@{}l|rr|rr|rr|rr|rr@{}}
  \toprule
  & \multicolumn{2}{c|}{oblique} & \multicolumn{8}{c}{axis-aligned} \\
  \midrule  
  & \multicolumn{2}{c|}{TAO} & \multicolumn{2}{c|}{TAO} & \multicolumn{2}{c|}{CART-R} & \multicolumn{2}{c|}{CART-P} & \multicolumn{2}{c}{GUIDE}\\
  Dataset&	$\Delta$ &	L&	$\Delta$&	L&	$\Delta$&	L&	$\Delta$&	L&	$\Delta$&	L\\
\midrule
  concrete&	8.0&	169.5&	11.2&	113.0&	11.8&	104.4&	11.2&	113.0&	9.0&	59.9\\
  airfoil&	8.0&	147.1&	15.0&	479.8&	15.6&	457.2&	15.0&	479.8&	10.6&	95.3\\
  abalone&	6.0&	58.6&	5.0&	12.8&	4.8&	11.0&	5.0&	12.8&	6.0&	18.1\\
  cpuact&	6.0&	52.7&	9.0&	57.2&	8.7&	52.8&	9.0&	57.2&	8.4&	21.8\\
  ailerons&	6.0&	60.2&	7.0&	15.0&	7.8&	66.6&	7.0&	15.0&	8.5&	66.1\\
  CT slice           &	7.0&	74.8&	36.0&	700.0&	30.0&	691.6&	36.0&	700.0&	12.7&	83.2\\
  YearPredictionMSD  &	8.0&	157.9&	12.0&	135.0&	11.8&	121.0&	12.0&	135.0&	10.3&	111.9\\
  \bottomrule
  \end{tabular}
  \caption{Average depths ($\Delta$) and average number of leaves (L) over 10 repetitions for regression datasets. TAO produces more compact and shallower trees which are easier to interpret.}
  \label{t:comp-regr-depth-leafno}
\end{table}

\subsection{Runtime}

It is commonly accepted that tree induction algorithms are fast. Especially, those which use greedy growing strategy like CART, C5.0, etc. We did not apply any parallel processing in our experiments and we observe the following:

\begin{itemize}

\item For small UCI datasets (like Balance Scale, Breast Cancer, etc.), all axis-aligned and oblique trees that we ran are quite fast (around $0.5$-$1.5$ seconds).
\item For relatively larger datasets (like MNIST, SensIT, etc.), we observe some fluctuations. In general, axis-aligned trees are still fast to train: for instance, CART and C5.0 took about 200-400 seconds to train on MNIST (pixels), whereas GUIDE and TAO axis-aligned took a little longer than that (about 1300-1500 seconds). As for oblique trees: GUIDE took extremely long time (about 26000 seconds for MNIST)and OC1 performs comparatively better (about 1800 seconds). Finally, TAO oblique decision tree optimization is much faster compare to other oblique tree training algorithms. It took about 1200-1400 seconds to train on MNIST.

\end{itemize}

\section{Discussion}

In this work, we demonstrate the performance comparison of some well-known decision tree algorithms along with the recently proposed TAO algorithm. We evaluate their performance on classification and regression tasks which are the main tasks in machine learning. Our experiments show that TAO not only performs better in accuracy but also provides smaller and more interpretable decision trees.  The reason for such performance is the way how TAO optimization works. Traditional algorithms greedily optimize the decision trees: at each step they split the data by using a single or collection of attributes that optimize some criterion (usually impurity). This approach has no guarantees towards the reduction of the desired loss (e.g. misclassification loss). Thus, in the end, the obtained trees usually do not generalize well and also are big in size. On the other hand, TAO 
instead of optimizing the impurity of a node at each step, optimizes the desired objective function over a decision tree of given structure and finds much better approximate optima than CART-type algorithms. This is done by employing alternating optimization over the nodes of a tree. This approach leads to better-optimized decision trees that not only generalize well but also have a smaller size. Empirical results presented in this paper show that TAO outperforms other non-greedy approaches as well (e.g. CO2, soft trees, etc.)


\begin{thebibliography}{15}
\providecommand{\natexlab}[1]{#1}
\providecommand{\url}[1]{\texttt{#1}}
\expandafter\ifx\csname urlstyle\endcsname\relax
  \providecommand{\doi}[1]{doi: #1}\else
  \providecommand{\doi}{doi: \begingroup \urlstyle{rm}\Url}\fi

\bibitem[Bertsimas and Dunn(2017)]{BertsimDunn17a}
D.~Bertsimas and J.~Dunn.
\newblock Optimal classification trees.
\newblock \emph{Machine Learning}, 106\penalty0 (7):\penalty0 1039--1082, July
  2017.

\bibitem[Breiman et~al.(1984)Breiman, Friedman, Olshen, and Stone]{Breiman_84a}
L.~J. Breiman, J.~H. Friedman, R.~A. Olshen, and C.~J. Stone.
\newblock \emph{Classification and Regression Trees}.
\newblock Wadsworth, Belmont, Calif., 1984.

\bibitem[Carreira-Perpi{\~n}{\'a}n and Tavallali(2018)]{CarreirTavall18a}
M.~{\'A}. Carreira-Perpi{\~n}{\'a}n and P.~Tavallali.
\newblock Alternating optimization of decision trees, with application to
  learning sparse oblique trees.
\newblock In S.~Bengio, H.~Wallach, H.~Larochelle, K.~Grauman, N.~Cesa-Bianchi,
  and R.~Garnett, editors, \emph{Advances in Neural Information Processing
  Systems (NEURIPS)}, volume~31, pages 1211--1221. MIT Press, Cambridge, MA,
  2018.

\bibitem[Fan et~al.(2008)Fan, Chang, Hsieh, Wang, and Lin]{Fan_08a}
R.-E. Fan, K.-W. Chang, C.-J. Hsieh, X.-R. Wang, and C.-J. Lin.
\newblock {LIBLINEAR}: A library for large linear classification.
\newblock \emph{J. Machine Learning Research}, 9:\penalty0 1871--1874, Aug.
  2008.

\bibitem[Hyafil and Rivest(1976)]{HyafilRivest76a}
L.~Hyafil and R.~L. Rivest.
\newblock Constructing optimal binary decision trees is {NP}-complete.
\newblock \emph{Information Processing Letters}, 5\penalty0 (1):\penalty0
  15--17, May 1976.

\bibitem[{LeCun} et~al.(1998){LeCun}, Bottou, Bengio, and Haffner]{Lecun_98a}
Y.~{LeCun}, L.~Bottou, Y.~Bengio, and P.~Haffner.
\newblock Gradient-based learning applied to document recognition.
\newblock \emph{Proc. IEEE}, 86\penalty0 (11):\penalty0 2278--2324, Nov. 1998.

\bibitem[Lichman(2013)]{Lichman13a}
M.~Lichman.
\newblock {UCI} machine learning repository.
\newblock \url{http://archive.ics.uci.edu/ml}, 2013.

\bibitem[Loh(2002)]{Loh02a}
W.-Y. Loh.
\newblock Regression trees with unbiased variable selection and interaction
  detection.
\newblock \emph{Statistica Sinica}, 12\penalty0 (2):\penalty0 361--386, Apr.
  2002.

\bibitem[Loh(2009)]{Loh09a}
W.-Y. Loh.
\newblock Improving the precision of classification trees.
\newblock \emph{Annals of Applied Statistics}, 3\penalty0 (4):\penalty0
  1710--1737, 2009.

\bibitem[Loh(2014)]{Loh14a}
W.-Y. Loh.
\newblock Fifty years of classification and regression trees.
\newblock \emph{International Statistical Review}, 82\penalty0 (3):\penalty0
  329--348 (with discussion, pp.~349--370), Dec. 2014.

\bibitem[Murthy et~al.(1993)Murthy, Kasif, Salzberg, and Beigel]{Murthy_93a}
S.~K. Murthy, S.~Kasif, S.~Salzberg, and R.~Beigel.
\newblock {OC1}: A randomized algorithm for building oblique decision trees.
\newblock In \emph{Proc. of the 11th National Conference on Artificial
  Intelligence (AAAI 1993)}, pages 322--327, Washington, DC, July~11--15 1993.

\bibitem[Norouzi et~al.(2015)Norouzi, Collins, Johnson, Fleet, and
  Kohli]{Norouz_15a}
M.~Norouzi, M.~Collins, M.~A. Johnson, D.~J. Fleet, and P.~Kohli.
\newblock Efficient non-greedy optimization of decision trees.
\newblock In C.~Cortes, N.~D. Lawrence, D.~D. Lee, M.~Sugiyama, and R.~Garnett,
  editors, \emph{Advances in Neural Information Processing Systems (NIPS)},
  volume~28, pages 1720--1728. MIT Press, Cambridge, MA, 2015.

\bibitem[Pedregosa et~al.(2011)Pedregosa, Varoquaux, Gramfort, Michel, Thirion,
  Grisel, Blondel, Prettenhofer, Weiss, Dubourg, Vanderplas, Passos,
  Cournapeau, Brucher, Perrot, and Duchesnay]{Pedreg_11a}
F.~Pedregosa, G.~Varoquaux, A.~Gramfort, V.~Michel, B.~Thirion, O.~Grisel,
  M.~Blondel, P.~Prettenhofer, R.~Weiss, V.~Dubourg, J.~Vanderplas, A.~Passos,
  D.~Cournapeau, M.~Brucher, M.~Perrot, and {\'E}.~Duchesnay.
\newblock Scikit-learn: Machine learning in {Python}.
\newblock \emph{J. Machine Learning Research}, 12:\penalty0 2825--2830, Oct.
  2011.
\newblock Available online at \url{https://scikit-learn.org}.

\bibitem[Quinlan(1993)]{Quinlan93a}
J.~R. Quinlan.
\newblock \emph{C4.5: Programs for Machine Learning}.
\newblock Morgan Kaufmann, 1993.

\bibitem[Therneau et~al.(2019)Therneau, Atkinson, and Ripley]{Therneau_19a}
T.~Therneau, B.~Atkinson, and B.~Ripley.
\newblock rpart: Recursive partitioning and regression trees.
\newblock R package version 4.1-15, Apr.~12 2019.
\newblock Available online at \url{https://cran.r-project.org/package=rpart}.

\end{thebibliography}

\end{document}